\documentclass[10pt,twocolumn,letterpaper]{article}

\usepackage{iccv}
\usepackage{times}
\usepackage{epsfig}
\usepackage{graphicx}
\usepackage{amsmath}
\usepackage{amssymb}
\usepackage{booktabs} 
\usepackage{multirow}
\usepackage{makecell}
\usepackage[mathscr]{eucal}

\usepackage{subfigure}
\usepackage{paralist}

\usepackage[pagebackref=true,breaklinks=true,letterpaper=true,colorlinks,citecolor=blue,linkcolor=blue,bookmarks=false]{hyperref}

\iccvfinalcopy 

\usepackage[table]{xcolor}

\def\iccvPaperID{2475} 

\ificcvfinal\fi

\newcommand*{\affaddr}[1]{#1} 
\newcommand*{\affmark}[1][*]{\textsuperscript{#1}}

\begin{document}
\makeatletter
\def\thanks#1{\protected@xdef\@thanks{\@thanks\protect\footnotetext{#1}}}
\makeatother
\title{CiteTracker: Correlating Image and Text for Visual Tracking}

\author{%
Xin Li\affmark[1,${\dag}$], Yuqing Huang\affmark[1,2,${\dag}$], Zhenyu He\affmark[2,$\ast$], Yaowei Wang\affmark[1,$\ast$], Huchuan Lu\affmark[3], and Ming-Hsuan Yang\affmark[4,5]\\
\affaddr{\affmark[1]Peng Cheng Laboratory}\quad
\affaddr{\affmark[2]Harbin Institute of Technology, Shenzhen}\\
\affaddr{\affmark[3]Dalian University of Technology} \quad 
\affaddr{\affmark[4]UC Merced}
\quad
\affaddr{\affmark[5]Yonsei University}\\
}
\thanks{$^\dag$ Equal contribution, $\ast$ corresponding author}
\maketitle
\ificcvfinal\thispagestyle{empty}\fi

\begin{abstract}
Existing visual tracking methods typically take an image patch as the reference of the target to perform tracking. 
However, a single image patch cannot provide a complete and precise concept of the target object as images are limited in their ability to abstract and can be ambiguous, which makes it difficult to track targets with drastic variations.
In this paper, we propose the CiteTracker to enhance target modeling and inference in visual tracking by connecting images and text.
Specifically, we develop a text generation module to convert the target image patch into a descriptive text containing its class and attribute information, providing a comprehensive reference point for the target.
In addition, a dynamic description module is designed to adapt to target variations for more effective target representation.
We then associate the target description and the search image using an attention-based correlation module to generate the correlated features for target state reference.
Extensive experiments on five diverse datasets are conducted to evaluate the proposed algorithm and the favorable performance against the state-of-the-art methods demonstrates the effectiveness of the proposed tracking method.
The source code and trained models will be released at \url{ https://github.com/NorahGreen/CiteTracker}.
\end{abstract}

\section{Introduction}

Visual object tracking aims to estimate the state (location and extent) of an arbitrary target in a video sequence based on a specified region of the target in the initial frame as a reference point.
It is challenging to locate a target that undergoes drastic appearance variations (\eg, changes in pose, illumination, or occlusions) using only one target image sample since the target appearances can be significantly different.
To successfully track the target with appearance changes, it is crucial to acquire a comprehensive representation of the target for establishing associations between the target exemplar and the target in test frames.

\begin{figure}[t]
\begin{center}
\includegraphics[width=0.9\linewidth]{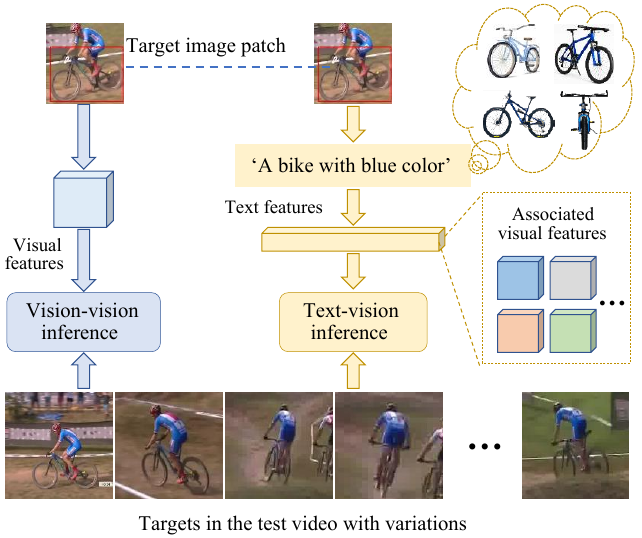}
\end{center}
   \caption{Comparison of the proposed algorithm and existing tracking methods in terms of target modeling and association. 
   The left and right parts depict the typical visual tracking framework and the proposed one, respectively.
   Our approach first generates a text description of the target object and then takes the feature of the text to estimate the target state in the test image, enabling a more comprehensive target modeling and association.}
\label{fig:motivation}
\end{figure}

Most existing deep trackers~\cite{SIAMESEFC, SiamRPN++, TrDiMP, TransT, Stark} learn an embedded feature space, where target samples with different appearances are still close to each other, to generate a robust representation to target variations. 
To build a more comprehensive target representation and better associate the target exemplar with the test target, several recent trackers~\cite{SimTrack, OSTrack, SwinTrack} perform interaction of the target template and search region in every block of their feature extraction backbone, achieving state-of-the-art tracking performance. 
However, these methods do not perform well when the target changes drastically or the given target exemplars are of low-quality.
%

The following issues arise when using an image patch as the target reference for tracking.
First, the visual representation of a target is insufficient to provide a comprehensive understanding for recognizing the target with appearance variations, since images are limited in their ability to abstract.
An image patch of a target only captures its appearance from a particular angle, but its shape, texture, and surface features can vary significantly when viewed from different angles, resulting in a completely different appearance that makes it difficult to track the target.
Second, as images can be ambiguous and open to interpretation, a random target image patch can mislead tracking models by causing them to overemphasize certain unstable appearance features and ignore the more essential and stable features of the target, resulting in drifting to the background and tracking failures.
For example, when tracking a circular object, the target patch may include a lot of the background, which causes the tracker to drift to the background.

We note that the human-created language signal provides a more abstract and precise concept of an object compared to the image signal, which has the potential to solve the aforementioned issues.
In addition, the study~\cite{CLIP} on connecting language and images shows that text 
and image features can be well-aligned and transferred to each other, allowing for using the advantages of both language and image signals for visual tracking.
Motivated by these insights, we study correlating text and images for visual tracking. 

%
In this paper, we propose a new tracking framework that uses an adaptive text description of the target as the reference point and correlates it with test image features to perform tracking, named as CiteTracker.
Specifically, we first develop a text generation model via prompt learning with a pre-defined open vocabulary including class and attribute labels, enabling generating the text description of the target based on a target image patch.
The generation model is built using the CLIP model as the baseline, which already connects text with rich image features.
To adapt to target variations over time, we develop a dynamic text feature model that generates adaptive text features along the change of the target.
Finally, we associate the features of the target text description with test image features to generate the correlated features for further target state estimation.
We conduct extensive experiments on a variety of public datasets including GOT-10K~\cite{GOT10k}, LaSOT, TrackingNet, OTB100, and TNL2K to evaluate the proposed algorithm.
The favorable performance against the state-of-the-art methods on all the datasets demonstrates the effectiveness of correlating images and text for visual tracking.

%

We make the following contributions in this paper:
\begin{compactitem}
    \item We propose a text-image correlation based tracking framework.
    We use a text description to provide a more comprehensive and precise concept of the target and correlate the text with the test image for inferring the target location, enabling a more powerful ability to handle the target variation issue.
    \item We develop an adaptive feature model for
    target descriptions to better adapt to target variations in test videos, contributing to more precise target features and more accurate tracking performance. 
    \item We achieve state-of-the-art performance on numerous tracking datasets.
    We conduct extensive experiments including ablation studies to demonstrate the effectiveness of the proposed method and the effect of every component.
\end{compactitem}

\begin{figure*}[!t]
	\centering
	\includegraphics[width=0.95\linewidth]{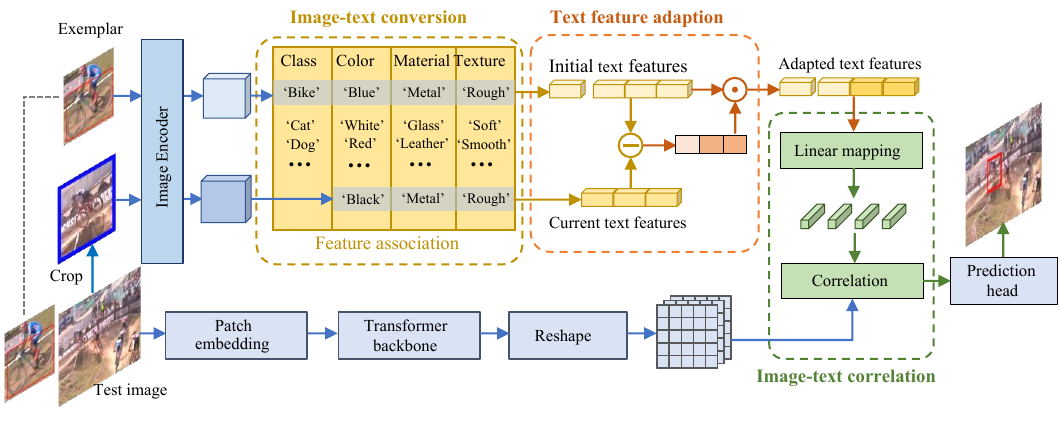}
	\caption{\textbf{Overall framework of the proposed CiteTracking algorithm.}
		It contains three modules: 1) the image-text conversion module, which generates text features of the target object based on an image exemplar; 2) the text feature adaption module, which adjusts the weights of attribute descriptions according to the current target state; 3) the image-text correlation module, which correlates the features of the target descriptions and the test image to generate correlated features for target state estimation.
	}
	\label{Fig:framework}
\end{figure*}

\section{Related Work}
We discuss closely related studies including deep visual tracking methods, language-based tracking approaches, and language-image association models.

\vspace{2mm}
\noindent\textbf{Deep visual-based trackers.}
Deep visual tracking methods can be broadly categorized as correlation-based or model-prediction-based, depending on how they model and infer the target during tracking.
The correlation-based trackers~\cite{SiamRPN, DaSiamRPN, SiamBAN, SiamRPN++, OCEAN, SiamR-CNN} associate the target exemplar with the test image in embedded feature space to generate the correlated features and estimate the target state upon the correlated features using classification and box-regression prediction heads.
As the correlated features show the similarity of every location of the test image to the target exemplar, their quality directly affects tracking performance.
By using more powerful features~\cite{transformer, VIT, DETR, SwinTrack}, the transformer-based trackers~\cite{TransT, TrDiMP, Stark} significantly improve the tracking performance of previous CNN-based trackers.
In addition, several recent transformer-based tracking methods~\cite{OSTrack, SimTrack, MixFormer} improve the association way by interacting the target exemplar and the test image in every transformer block for a more comprehensive correlation, achieving state-of-the-art performance.
The model-prediction-based trackers~\cite{ATOM, DiMP, PrDiMP, ToMP} learn to generate a classifier model based on the given target image patch and the backgrounds in the initial frame.
These methods are effective at distinguishing the target from the background by specifically learning the difference between them, resulting in a strong discriminative ability.
Different from the above deep visual trackers, the proposed method models the target also with language and infers the target via the language-image correlation, which enables a more comprehensive target modeling and robust target inference.

\vspace{2mm}
\noindent\textbf{Language-based trackers.}
Several methods~\cite{TNLS, TNL2K, SNLT, CANLS, VLT} explore utilizing language signals to facilitate visual object tracking.
Some of them use the language signal as an additional cue and combine it with the commonly used visual cue to compute the final tracking result.
The SNLT tracker~\cite{SNLT} first exploits visual and language descriptions individually to predict the target state and then dynamically aggregates these predictions for generating the final tracking result.
In \cite{TNL2K}, Wang \etal propose an adaptive switch-based tracker that switches to a visual grounding module when the target is lost and switches back to a visual tracking module when the target is found, ensuring robust and precise tracking.
Another type of method focuses on integrating the visual and textual signals to get an enhanced representation for visual tracking.
The CapsuleTNL~\cite{CANLS} tracker develops a visual-textual routing module and a textual-visual routing module to promote the relationships within feature embedding space of query-to-frame and frame-to-query for object tracking.
In \cite{VLT}, a modality mixer module is developed to learn unified-adaptive visual-language representations for robust vision-language tracking.
Despite also utilizing both language and visual information for tracking, our method differs significantly from the above methods in terms of how to generate text descriptions of the target and associate them with the search image to perform tracking.
The method proposed develops a CLIP-based model to generate text descriptions from a target image example, which eliminates the need for language annotations and expands the range of potential applications.
In addition, we design a dynamic feature reweighting module that adjusts language features based on target appearance changes, leading to more accurate tracking performance.

\vspace{2mm}
\noindent\textbf{Vision-language models.}
%
Recently, the CLIP model~\cite{CLIP}, trained on a large and diverse dataset of images and their associated captions, maps images and their corresponding text descriptions into a shared feature space, where the similarity between the two modalities can be measured.
This shared feature space allows the model to perform various tasks, such as image generation~\cite{DALLE}, few-shot learning~\cite{MMfewshot}, and image captioning~\cite{ImageCap}. 
CLIP-based methods achieve state-of-the-art performance on several benchmarks for these tasks, demonstrating their ability to generalize to new domains and languages.
Based on the CLIP model, we develop a dynamic text feature generation module to enable more comprehensive target modeling with more accurate and informative representations for visual tracking.
In addition, as text and image features are well aligned in the CLIP model, we correlate the text features of the target with the search image features for reasoning about the location of the target, achieving more robust tracking performance.

\section{Proposed Algorithm}
The goal of our approach is to construct a robust association between a given target image patch and a search image in a tracking sequence by formulating it as an image-text correlation allowing a more comprehensive understanding of the target state, which helps cope with various appearance variations of the target object.
To this end, our CiteTracker first generates text features of the target based on the given target image patch via the proposed image-text conversion module, then adjusts the text features according to the latest state of the target, and finally associates the features of the text and the search image for robust tracking.

\begin{figure}[t]
\begin{center}
\includegraphics[width=0.82\linewidth]{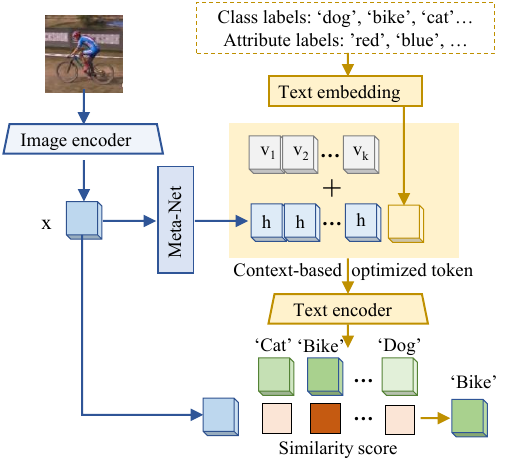}
\end{center}
   \caption{\textbf{Structure of the image-text conversion module.} It takes a target image and pre-defined class \ attribute text vocabularies as input and outputs text descriptions (features) of the target.}
\label{fig:finetune}
\end{figure}

\subsection{Overall Framework}
Figure~\ref{Fig:framework} shows the overall framework of the CiteTracker, consisting of three core modules: an image-text conversion module, a text feature adaption module, and an image-text correlation part.

Taken as input an exemplar image and a search image in a test sequence, our approach processes them with a text branch (the upper part in Figure~\ref{Fig:framework}) and a visual branch (the lower part in Figure~\ref{Fig:framework}).  
The text branch first uses an image encoder to extract the visual features of the given exemplar image and a target image patch cropped from the test image at the target location in the previous frame.
Then, it converts the visual features of the target to text features with the image-text conversion module and adjusts the text features using the text feature adaption module based on the difference between the text features of the initial target state and the current target state.
The visual branch adopts the same processing flow as OSTrack~\cite{OSTrack} that takes both the exemplar and search images as input and outputs a feature map of the test image.
Finally, the image-text correlation component associates the outputs of the text and visual branches to generate the correlated features for target state prediction via a commonly used prediction head~\cite{Stark}.

\subsection{Image-Text Conversion}
To generate the text feature of a tracking target from a given image exemplar, we construct an image-text conversion model that connects images and text based on the CLIP model~\cite{CLIP} via prompt learning.

\begin{figure}[t]
\begin{center}

    \subfigure[Consistency of predictions on the GOT-10K training dataset.]{
        \begin{minipage}[b]{0.85\linewidth}
        \label{fig:consis-a}
             \includegraphics[width=1\linewidth]{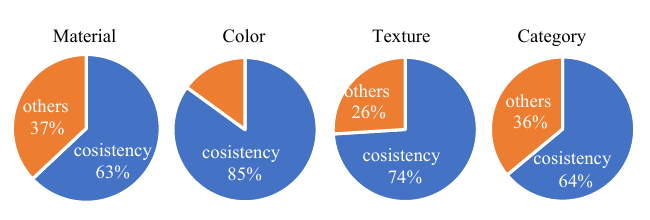}
    \end{minipage}
    \vspace{-4mm}
    }
    \subfigure[Temporal consistency on a video sequence.]{
        \begin{minipage}[b]{0.85\linewidth}
        \label{fig:consis-b}
             \includegraphics[width=1\linewidth]{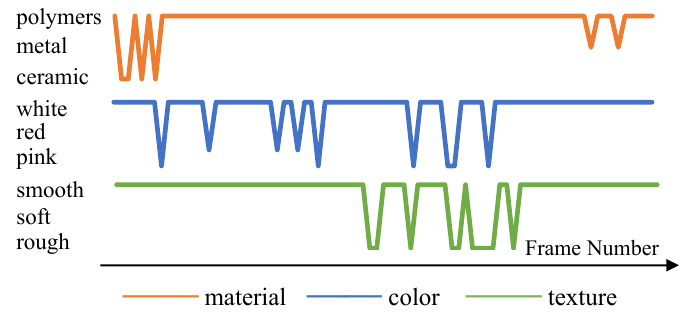}
        \end{minipage}
    }
    \vspace{-2mm}
\end{center}
   \caption{\textbf{Consistency of the predicted descriptions in terms of target category and the selected attributes on tracking videos.}
   }
\label{fig:consistency}
\end{figure}

\vspace{2mm}
\noindent\textbf{Image-text association learning.}
Figure~\ref{fig:finetune} shows the structure of the image-text conversion model.
It takes the target image and vocabulary of object categories and attributes as input.
The target image is processed by the image encoder of a CLIP model to generate the image feature $x$ and $x$ is then fed into a lightweight neural network $h_\theta(\cdot)$ (Meta-Net) to generate target tokens $h_\theta(x)$ that contain the target information.
The input vocabulary is processed by a text embedding module to generate the word embeddings $\it{c_i}$.
In addition, $K$ learnable vectors ${\upsilon_1,\upsilon_2,...,\upsilon_K}$, where $\upsilon_i$ have the same dimensions as $\it{c_i}$, are introduced as prompt tokens for a specific prediction task.
Given the target tokens $h_\theta(x)$ and the prompt tokens ${\upsilon_1,\upsilon_2,...,\upsilon_K}$, each context-based optimized token can be obtained by 
$\upsilon_k (x)$ =  $\upsilon_k$ + $h_\theta(x)$, where $k\in\{1,2,...,K\}$.
The prompt for the $i$-th class label is thus conditioned on the image features, i.e. $\it{m_i}(x)$ = $\{\upsilon_1(x),\upsilon_2(x),...,\upsilon_K(x),\it{c_i}\}$. 
Let $t(\cdot)$ denote the original CLIP text encoder and the prediction probability of the $i$-th class label is computed as
\begin{equation}
\label{equation_clip}
    p(c_i|x)=\frac{\exp(S_{im}(x, \it{t(m_i(x)))/\tau)}}{ {\textstyle \sum_{j=1}^{N}}\exp(S_{im}(x, \it{t(m_j(x)))/\tau)} },
\end{equation}
where $S_{im}(\cdot, \cdot)$ computes the cosine similarity score, $\tau$ is a learned temperature parameter, and $N$ is the number of class labels. 
The target description is predicted as the label corresponding to the max probability computed with Equation~\ref{equation_clip}.
In this work, we implement the Meta-Net using a two Linear-ReLU-Linear structure with the hidden layer reducing the input dimension by 16 times. 

\vspace{2mm}
\noindent\textbf{Tracking-related vocabulary construction.} 
In order to accurately describe tracking targets, we choose 80 category labels in the MS COCO~\cite{COCO} dataset as the category vocabulary, which contains the most frequently occurring objects in daily life. 
In addition, we select three kinds of object attributes including color, texture, and material from the OVAD~\cite{OVAD} dataset to caption detailed target states.
We evaluate the consistency of the predicted descriptions in terms of the class and attribute labels on the GOT-10k dataset.
Figure~\ref{fig:consis-a} shows the proportions of cases where the predicted results are consistent and Figure~\ref{fig:consis-b} presents the predicted values of a target object in video frames.
They demonstrate that the predicted text descriptions of tracking objects in terms of class and attribute values are consistent in video sequences, which can be used as features for target localization.

\subsection{Dynamic Text Feature Generation}

\vspace{2mm}
In a video, the category of tracking target remains consistent but its states may vary.
Therefore, we divide text feature generation into category feature generation and  attribute feature generation.
For category feature $T_c$ , let $T_i$ is the text feature of $i$-th class label generated by the CLIP text encoder, $T_c$ can be computed as
\begin{equation}
       T_c = \sum_{i=1}^{N} p_i * T_i,
\end{equation}
where $p_i$ is the prediction probability of every category label using Equation~\ref{equation_clip}. For each attribute feature $T_a$, which has the highest prediction probability, can be computed as
\begin{equation}
\begin{aligned}
       index &= argmax(p_i), i\in(1,N), \\
       T_{a} &= T_{index}.
\end{aligned}
\label{att_feature}
\end{equation}

As the attribute values of a tracking target may change, we adjust the weights of different attribute features based on their changes. 
The changes in terms of color, material, and texture, denoted as $D_{color}$, $D_{material}$, and $D_{texture}$, respectively, are computed as
\begin{equation}
\begin{aligned}
       D_{color} &= \left |  R_{color} - S_{color} \right |,\\
       D_{material} &= \left | R_{material} - S_{material} \right |, \\
       D_{texture} &= \left |R_{texture} - S_{texture} \right |,
\end{aligned}
\end{equation}
where $R_{attribute}$ and $S_{attribute}$ represent the probabilities of the reference target and the current test target to be with a specific attribute value computed using Equation~\ref{equation_clip}.
The lower the $R_{attribute}$ value,  the more similar the target and the search image are on that attribute.
Therefore, the attention weights for different attributes are formulated as:
 \begin{equation}
 \label{att-weight}
       W_{att} = 
       Softmax(-D_{color}, -D_{material}, -D_{texture}).
\end{equation}
After that, the dynamic text features for different attributes are adjusted as
 \begin{equation}
 \label{dy-att-weight}
       T_{att} = W_{att} * T_{a},
\end{equation}
where $T_{a}$ is the text feature generated by using Equation~\ref{att_feature}. 

\subsection{Image-Text Correlation}
The joint visual features $V \in \mathbb{R}^{\it{H \times W \times C}}$ of the target and the search image are extracted by the Vision Transformer (ViT-base)~\cite{transformer} model pre-trained with the MAE~\cite{MAE} method. 
The text features $T \in \mathbb{R}^{\it{1 \times 1 \times C_{T}}}$ are adapted by a linear layer to align with the visual features in the channel dimension.
Then the correlation between these two kinds of features is achieved by a convolution operation where the text features $T' \in \mathbb{R}^{\it{1 \times 1 \times C }}$ are used as the kernel weights.
The correlated features between the image features and all the text features are added up as the final correlated features for state prediction, which are computed as
\begin{equation}
\begin{aligned}
    & C_{orr}(V, T_c, T_{co}, T_m, T_t) = \\
    & (1 + {\rm L_{c}}(T_c) + {\rm L_{co}}(T_{co}) + {\rm L_{m}}(T_m) + {\rm L_{t}}(T_t))\odot V,
\end{aligned}
\end{equation}
where $\odot$ denotes the convolutional operation, ${\rm L_{att}}$ is a linear projection layer for channel adaptation,  $T_c$ denotes the category feature, while $T_{co}$, $T_m$, and $T_t$ represent the dynamic color, material, and texture feature, respectively. 
%

\vspace{2mm}
\subsection{State Estimation and Training Objective}
\noindent\textbf{State estimation.} 
Based on the correlated features generated by the image-text correlation, our CiteTracker estimates the target state via a commonly used prediction head~\cite{OSTrack} comprising 4 stacked Conv-BN-ReLU layers.
The prediction head outputs a classification score map $C$, offset maps $O$ for compensating for reduced resolution, and size maps $B$. 
Then, the target state is computed as
\begin{equation}
\begin{aligned}
(x,y,w,h)=(x_c+O_x, y_c+O_y, B_w, B_h),
\end{aligned}
\end{equation}
where $(x_c, y_c)$ is the target center computed as $(x_c, y_c)=argmax_{(x,y)} C_{xy}$, $(O_x,O_y)$ denotes the shifts to $(x_c, y_c)$ from $O$, and $(B_w, B_h)$ is the predicted box size from $B$.

\vspace{1mm}
\noindent\textbf{Training objective.}  
We adopt a similar training process as that of OSTrack~\cite{OSTrack}, which trains the three tasks jointly. 
We use the weighted focal loss~\cite{FocalLoss}, $\it{l_1}$ Loss, and the GIoU~\cite{GIoU} loss to train the classification, offset, and box size branches, respectively.
The overall loss function is defined as
\begin{equation}
\begin{aligned}
\label {equ-loss-loc} 
L=L_{cls} + \lambda_{iou}L_{iou} + \lambda_{L1}L_1, \
\end{aligned}
\end{equation}
where $\lambda_{iou}$ = 2 and $\lambda_{L1}$ = 5 are used in our experiments.


\begin{table*}
\rowcolors{2}{gray!25}{white}
\begin{center}
 \caption{\textbf{State-of-the-art comparisons on the datasets of TNL2K, LaSOT, TrackingNet, and GOT-10k.} The best two results are shown in \textcolor[rgb]{1,0,0}{red} and \textcolor[rgb]{0,0,1}{blue} color. Our approach performs favorably against the state-of-the-art methods on all datasets.}
 \vspace{1mm}
 \label{tab:sota-eval}
 \small
\begin{tabular}{ll*{11}{c}}
\toprule 
 \multirow{2}*{Method} 
& \multicolumn{2}{c}{TNL2K~\cite{TNL2K}} & \multicolumn{3}{c}{LaSOT~\cite{LaSOT}} & \multicolumn{3}{c}{TrackingNet~\cite{TrackingNet}} & \multicolumn{3}{c}{GOT-10k~\cite{GOT10k}}

\\\cmidrule(lr){2-3}\cmidrule(lr){4-6}\cmidrule(lr){7-9}\cmidrule(lr){10-12}
   & P & SUC & AUC & P$_{Norm}$ & P    & AUC  & P$_{Norm}$ & P  & AO  & SR$_{0.75}$ & SR$_{0.5}$\\\midrule
  SiamFC~\cite{SIAMESEFC} & 28.6 & 29.5  & 33.6 & 42.0 & 33.9 & 57.1 & 66.3 & 53.3 & 34.8 & 9.8 & 35.3\\

 RPN++~\cite{SiamRPN++}  & 41.2 & 41.3  & 49.6 & 56.9 & 49.1 & 73.3 & 80.0 & 69.4 & 51.7 & 32.5 & 61.6\\

 Ocean~\cite{OCEAN}  & 37.7 & 38.4  & 56.0 & 65.1 & 56.6 & - & - & - & 61.1 & 47.3 & 72.1 \\

 TransT~\cite{TransT}  & 51.7 & 50.7  & 64.9 & 73.8 & 69.0 & 81.4 & 86.7 & 80.3 & 67.1 & 60.9 & 76.8\\

 KeepTrack~\cite{KeepTrack}  & - & -  & 67.1 & 77.2 & 70.2 & - & - & - & - & - & - \\

 STARK~\cite{Stark}  & - & - & 67.1 & 77.0 & - & 82.0 & 86.9 & - & 68.8 & 64.1 & 78.1 \\

 OSTrack~\cite{OSTrack} & - & \textcolor[rgb]{0,0,1}{55.9}   & \textcolor[rgb]{0,0,1}{71.1} & \textcolor[rgb]{1,0,0}{81.1} & \textcolor[rgb]{1,0,0}{77.6} & 83.9 & 88.5 & \textcolor[rgb]{0,0,1}{83.2} & \textcolor[rgb]{0,0,1}{73.7} & \textcolor[rgb]{0,0,1}{70.8} & \textcolor[rgb]{0,0,1}{83.2} \\

 SimTrack~\cite{SimTrack} & 55.7 & 55.6  & 70.5 & 79.7 & - & 83.4 & 87.4  & - & - & - & -\\ %

 MixFormer~\cite{MixFormer} & - & -  & 70.1 & \textcolor[rgb]{0,0,1}{79.9} & 76.3 & 83.9 & \textcolor[rgb]{0,0,1}{88.9} & 83.1 & 71.2 & 65.8 & 79.9\\

 AiATrack~\cite{AiATrack} & - & - & 69.0 & 79.4 & 73.8 & 82.7 & 87.8 & 80.4 & 69.6 & 63.2 & 80.0\\

 SwinTrack~\cite{SwinTrack} & \textcolor[rgb]{0,0,1}{57.1} & \textcolor[rgb]{0,0,1}{55.9}  & \textcolor[rgb]{1,0,0}{71.3} & - & \textcolor[rgb]{0,0,1}{76.5} & \textcolor[rgb]{0,0,1}{84.0} & - & 82.8 & 72.4 & 67.8 & 80.5\\\midrule

 SNLT~\cite{SNLT} & 41.9 & 27.6  & 54.0 & - & 57.6 & - & - & - & 43.3 & 22.1 & 50.6\\
 VLT~\cite{VLT} & 53.3 & 53.1 & 67.3 & - & 72.1 & - & - & - & 69.4 & 64.5 & 81.1\\

\midrule
 Ours  & \textcolor[rgb]{1,0,0}{59.6} & \textcolor[rgb]{1,0,0}{57.7} & 69.7  & 78.6 & 75.7 & \textcolor[rgb]{1,0,0}{84.5} & \textcolor[rgb]{1,0,0}{89.0} & \textcolor[rgb]{1,0,0}{84.2} & \textcolor[rgb]{1,0,0}{74.7} & \textcolor[rgb]{1,0,0}{73.0} & \textcolor[rgb]{1,0,0}{84.3}
\\\bottomrule
\end{tabular}
\end{center}
\end{table*}

\section{Experiments}
In this section, we present the experimental results of the proposed CiteTracker.
We first show the overall performance on four large-scale datasets with comparisons against the state-of-the-art trackers.
We then investigate the contribution of each component with an exhaustive ablation study.
A robustness evaluation is conducted to study the robustness of our tracker to initialization. 
Finally, the visualized results on a number of challenging sequences are given for providing a comprehensive qualitative analysis.

\subsection{Implementation Details}
Our experiments are conducted with 4 NVIDIA Tesla V100 GPUs. 
We adopt the Vision Transformer (ViT-base)~\cite{transformer} model pre-trained using the MAE~\cite{MAE} method as the backbone for extracting visual features. 
We use the fine-tuned version of the CLIP model~\cite{CLIP} as the backbone to construct the proposed image-text conversion model.
We crop the search image that is 4 times the area of the target box from the test frame and resize it to a resolution of ${384}\times{384}$ pixels. 
While crop only 2 times of that from the reference frame and resize it to ${192}\times{192}$ pixels. 
The open vocabulary class labels and attribute labels are derived from the MS COCO~\cite{COCO} dataset and OVAD~\cite{OVAD} dataset.
We train our CiteTracker on the training splits of the TrackingNet~\cite{TrackingNet}, COCO2017~\cite{COCO}, LaSOT~\cite{LaSOT}, and GOT-10K~\cite{GOT10k} datasets, except for the evaluation on GOT-10K, where CiteTracker is only trained on the GOT-10K training set.


\subsection{State-of-the-art Comparison}

We compare our tracker with the state-of-the-art methods on four diverse datasets including TNL2K, LaSOT, TrackingNet, and GOT-10K. 
Table~\ref{tab:sota-eval} shows the results.

\vspace{1mm}
\noindent \textbf{TNL2K~\cite{TNL2K}.} TNL2K is a benchmark designed for evaluating natural language-based tracking algorithms. 
The benchmark introduces two new challenges, \ie adversarial samples and modality switching, which makes it a robust benchmark for tracking algorithm assessment. 
Although the benchmark provides both bounding boxes and language descriptions, we only use the bounding box for evaluation. 
Our approach achieves the best performance compared to the state-of-the-art methods including the language-based VLT tracker.
Compared to the second-best tracker OSTrack~\cite{OSTrack}, the proposed method improves the performance by gains of 1.8$\%$ and 2.5$\%$ in terms of success rate (SUC) and precision, respectively.
The favorable performance demonstrates the promising potential of our tracker to deal with adversarial samples and modality switch problems, which benefits from the use of text descriptions to model and inference the tracking target.

\vspace{1mm}
\noindent \textbf{LaSOT~\cite{LaSOT}.} LaSOT is a high-quality long-term single object tracking benchmark, with an average video length of more than 2,500 frames.
%
Although our method does not employ any updating mechanisms which play a critical role in long-term tracking, it still achieved a result close to the best method SwinTrack.
The proposed CiteTracker focuses on handling drastic target variations by formulating target inference as a robust image-text correlation.

 \vspace{1mm}
\noindent 
\textbf{TrackingNet~\cite{TrackingNet}.} TrackingNet is a large-scale short-term benchmark for object tracking in the wild, which contains 511 testing videos that sequester the ground truth annotation. 
Table~\ref{tab:sota-eval} shows the performance on the TrackingNet dataset. 
Our tracker achieves 84.4$\%$ in area-under-the-curve (AUC), surpassing all previously released trackers. 
It depicts that our tracker is highly competitive in tracking short-term scenarios in the wild with various changes.

\vspace{1mm}
\noindent \textbf{GOT-10k~\cite{GOT10k}.} GOT-10k is a large-scale tracking dataset that contains over 560 classes of moving objects and 87 motion patterns, emphasizing class agnosticism in the test set. 
The ground truths of the test set are withheld and we use the test platform provided by the authors to evaluate our results. 
We follow the one-shot protocol training rule that the tracker is trained only on the training set of GOT-10k. 
As shown in Table~\ref{tab:sota-eval}, our tracker improves all metrics, e.g. 1.5$\%$ in AUC score compared with OSTrack~\cite{OSTrack} and SwinTrack~\cite{SwinTrack}.
The good performance shows that our tracker has a good generalization ability to track class-agnostic targets. 
We attribute this to the proposed robust target modeling approach using text descriptions.

\vspace{1mm}
\noindent\textbf{Comparisons with Vision-Language trackers.}
In addition to the comparisons with the SOTA visual trackers, we compare the proposed method with the SOTA vision-language trackers to verify the effectiveness of the description generation ability. 
Our tracker improves the tracking performance in all benchmarks by a large margin, \eg 5$\%$ in the success rate of TrackingNet benchmark and 5.3$\%$ in that of GOT-10k compared with the recently published vision-language tracker VLT~\cite{VLT}. 
Although we do not use the manually annotated text description, the proposed method with a description generation module can still achieve considerable tracking performance.

\newcommand{\tabincell}[2]{\begin{tabular}{@{}#1@{}}#2\end{tabular}}
\begin{table}
\begin{center}   
 \caption{\textbf{Ablation study of the proposed algorithm on the OTB, GOT-10k, and TNL2K datasets. }
 The best results in each part of the table are marked in \textbf{bold}.}
 \vspace{1mm}
  \label{tab:ablation}
 \footnotesize
\resizebox{1.0\linewidth}{!}{
\begin{tabular}{*{8}{c}}
\toprule 
& & \tabincell{c}{Base\\model}    & \tabincell{c}{Lang\\Traker}  &\tabincell{c}{w/o\\FT}   &\tabincell{c}{w/o\\DDG}   &\tabincell{c}{w/o\\attr.} &  \tabincell{c}{Cite\\Tracker} \\\midrule
 \multirow{3}*{OTB} & AUC & 67.5 & 68.1 & 69.1 & 69.3 & 69.3  & \textbf{69.6}\\
 &P & 88.8 & 90.2 & 90.4 & 90.6 & 90.6  & \textbf{92.2}\\
 &NP  & 82.7 & 83.5 & 84.1 & 84.2 & 84.4 & \textbf{85.1} \\\midrule
  \multirow{3}*{GOT-10k} & AO & 72.9 & - & 71.4 & 74.5 & 74.2  & \textbf{74.7}\\
 &SR$_{0.75}$ & 70.2 & - & 68.8 & 72.5 & 72.4 & \textbf{73.0} \\
 &SR$_{0.5}$ & 82.4 & - & 80.4 & 83.8 & 83.5 & \textbf{84.3} \\\midrule
  \multirow{2}*{TNL2K} & P & 57.0 & 58.8 & 57.5 & 59.3 & 59.1  & \textbf{59.6}\\
 &SUC & 55.9 & 57.1 & 56.1 & 57.6 & 57.4  & \textbf{57.7}\\
\bottomrule
\end{tabular}}
\end{center}
\end{table}

\subsection{Ablation Study}
To evaluate the contribution of each component in our tracker, we conduct the ablation studies with six variants of the CiteTracker:

\noindent\textbf{Base visual model}, which only employs the backbone to extract the joint visual features of the target and test images, and a prediction head to predict the final tracking results. Herein, the prediction head is constructed on the feature maps of the joint visual features.

\noindent\textbf{LangTraker}, which uses manually annotated target descriptions to track. 
It extracts description features by the CLIP text encoder and performs a correlation between the extracted description features and visual features obtained from the backbone to acquire the associated features. 

\noindent\textbf{W/O attribute (attr.)}, which only generates category descriptions from the template frame using the image-text conversion model, and then correlates these descriptions with visual features extracted from the backbone to obtain associated features.

\noindent\textbf{W/O dynamic description generation (DDG)}, which extracts category and attribute descriptions only from the template frame using the image-text conversion model. 

\noindent\textbf{W/O fine-tune (FT)}, which employs the original CLIP model to extract category descriptions and attribute descriptions for the tracking targets.

\noindent\textbf{CiteTracker}, our intact model uses an image-text conversion model to obtain category and attribute descriptions for both the template and search frames. Then, these descriptions are correlated with visual features to generate the correlated features for target state estimation.

Table \ref{tab:ablation} presents the experimental results of these variants on the OTB2015, GOT-10K, and TNL2K datasets. The manually annotated target descriptions of the OTB dataset are from the OTB-lang dataset~\cite{OTB-lang}.

\vspace{1mm}
\noindent\textbf{Effect of the vision and text feature correlation.} The performance gap between the base model and LangTraker clearly demonstrates the advantages of correlating visual and text-based features for tracking.

\vspace{1mm}
\noindent\textbf{Effect of the prompt-tuning on the CLIP model.}
With the prompt-tuning process, CiteTracker achieves performance gains of 0.5$\%$ and 3.3$\%$ in AUC on OTB2015 and GOT-10K, while 1.6$\%$ in SUC on TNL2K, respectively. These improvements validate the benefits of the prompt-tuning of the CLIP model, which generates a more robust representation by exploiting the content-based optimization tokens.

\vspace{1mm}
\noindent\textbf{Effect of using the attribute description.} Without using the attribute description (w/o attr.), CiteTracker decreases by 1.6$\%$ and 0.5$\%$ in precision on OTB2015 and TNL2K respectively. It validates the superiority of using the attribute description to model tracking objects.

\vspace{1mm}
\noindent\textbf{Effect of the dynamic text-feature generation module.} By comparing our CiteTracker with w/o DDG, it is clear that the proposed dynamic text-feature generation module improves tracking performance by 0.3$\%$, 0.2$\%$, and 0.3$\%$ in terms of AUC on OTB2015, GOT-10K, and TNL2K, respectively. This mechanism successfully enables the tracker to focus more on the differences between the reference and search frames, leading to improved results.

\begin{table}[!t]
\caption{\textbf{Robustness evaluation against the OSTrack method on the OTB dataset.} Our CiteTracker is more robust to initialization compared to the OSTrack method.}\label{tab:robustness}
\vspace{1mm}
\centering
 \resizebox{1.0\linewidth}{!}{
\begin{tabular}{crrrr}
\toprule
   AUC (\%) & TRE & TRE-worst & SRE-shift & SRE-scale  \\
   \midrule
   OSTrack & 69.3 & 57.5 & 45.9 & 63.1  \\
   Ours & (+0.9) 70.2 & (+3.6) 61.1 & (+1.4) 47.3 & (+2.7) 65.8  \\
   \bottomrule
\end{tabular}
}
\end{table}

\begin{figure*}[!t]
\centering
\includegraphics[width=0.98\linewidth]{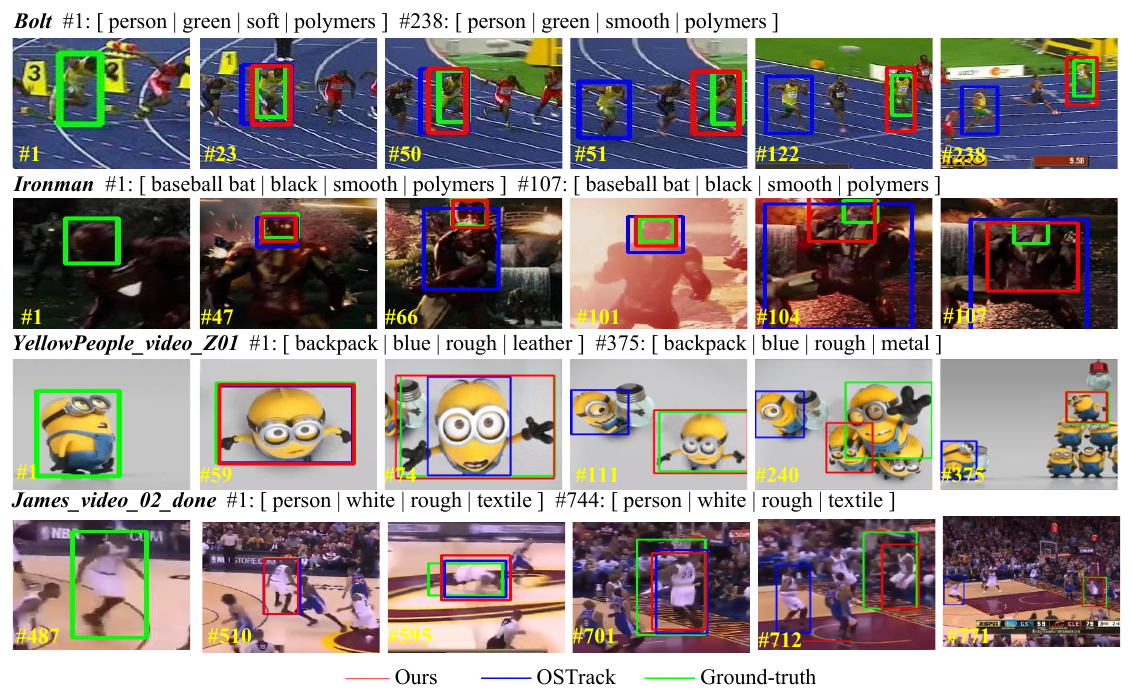}
 \caption{\textbf{Visualized results of the proposed algorithm and the OSTrack method on four challenging sequences with drastic changes.}
 It shows that our CiteTracker performs well with the aid of the generated text descriptions (shown above each row of pictures), while the OStrack method with solely visual cues struggles with these sequences.
 }
\label{Fig:case}
\end{figure*}

\subsection{Robustness Evaluation}
We evaluate the robustness of the proposed method on the OTB dataset by adopting the temporal robustness evaluation (TRE) and the spatial robustness evaluation (SRE)~\cite{OTB2013}.
We additionally report the worst score (TRE-worst) among all sequence segments, which measures the robustness of a tracker to bad initial target samples.
SRE-shit and SRE-scale denote the evaluations using shifted ground truth and scaled ground truth, respectively.
Table~\ref{tab:robustness} shows that the proposed tracker achieves more robust performance compared to OStrack, especially in terms of TRE-worst with a gain of 3.6\%, which demonstrates that our approach performs significantly well in the cases when the initial target examples are with very poor quality.

\subsection{Qualitative Study}
To obtain more insights from our proposed tracking algorithm, we visualize the tracking results of several challenging sequences compared with OSTrack.

The $\it{Bolt}$ sequence is characterized by a swiftly moving target and adversarial examples that closely resemble the reference target. Our tracking algorithm performs accurately in tracking the target, whereas OSTrack fails to track the target at the 51st frame. 
In the $\it{Ironman}$ series, our tracker accurately tracks the target despite significant lighting changes, whereas OSTrack does not. 
In addition, our CiteTracker accurately locates the target and distinguishes it from similar distractors, even in the presence of adversarial samples and target appearance variations in the $\it{YellowPeople}$ sequence. 
Despite frequent changes in viewpoint in the $\it{James}$ sequence, our tracking algorithm still performs well.

Figure~\ref{Fig:case} additionally shows the generated text descriptions of the targets for each sequence, including category, color, material, and texture. 
Most descriptions of the same target are consistent in different frames with drastic changes, which demonstrates the robustness of textual descriptions for tracking. 
The predicted category may differ from the real object class due to the limited 80 categories from the COCO dataset used for training, but it remains consistent in most frames of a video (is also supported by the statistical results in Figure~\ref{fig:consistency}), benefiting target identification and localization.

\section{Conclusions}
In this work, we present the CiteTracker that performs target modeling and target state inference in a more robust and accurate way by associating images and text.
Specifically, the proposed algorithm first constructs an image-text conversion model to generate text-description features of the target from a given target image, enabling a more abstract and accurate target representation.
In addition, we develop a text feature adaption model to generate dynamic text features and an image-text correlation to associate the target text and the search image for further target state prediction.
Qualitative and quantitative evaluations demonstrate that our approach performs favorably against the state-of-the-art methods, which suggests that incorporating language signals into visual tracking has a notable effect on improving tracking performance.

\section{Acknowledgments}
\label{sec:Acknowledgments}
The paper is supported by the National Natural Science Foundation of China (62002241, U20B2052, 62172126), and the Shenzhen Research Council (No. JCYJ20210324120202006).

{\small
\bibliographystyle{ieee_fullname}
\bibliography{tracking}
}

\end{document}